\documentclass{article}
\usepackage{ijcai19}

\usepackage{times}
\usepackage{soul}
\usepackage{url}
\usepackage[hidelinks]{hyperref}
\usepackage[utf8]{inputenc}
\usepackage[small]{caption}
\usepackage{graphicx}
\usepackage{amsmath}
\usepackage{amsfonts} 
\usepackage{amssymb} 
\usepackage{enumerate}
\usepackage{enumitem}
\usepackage{booktabs}
\usepackage{multirow}
\title{Machine Learning vs Statistical Methods for Time Series Forecasting: \\Size Matters}

\author{
Vitor Cerqueira$^{1,2}$\footnote{Contact Author}\and
Luis Torgo$^{1,2,3}$\And
Carlos Soares$^{1,2}$\\
\affiliations
$^1$INESC TEC, Porto, Portugal\\
$^2$University of Porto\\
$^3$Dalhousie University\\
\emails
vitor.cerqueira@fe.up.pt,
ltorgo@dal.ca,
csoares@fe.up.pt
}

\begin{document}

\maketitle

\begin{abstract}
Time series forecasting is one of the most active research topics. Machine learning methods have been increasingly adopted to solve these predictive tasks. However, in a recent work, evidence was shown that these approaches systematically present a lower predictive performance relative to simple statistical methods. In this work, we counter these results. We show that these are only valid under an extremely low sample size. Using a learning curve method, our results suggest that machine learning methods improve their relative predictive performance as the sample size grows. The R code to reproduce all of our experiments is available at \url{https://github.com/vcerqueira/MLforForecasting}.
\end{abstract}

\section{Introduction} 

Machine learning is a subset of the field of artificial intelligence, which is devoted to developing algorithms that automatically learn from data \cite{michalski2013machine}. This area has been at the centre of important advances in science and technology. This includes problems involving forecasting, such as in the domains of energy \cite{voyant2017machine}, healthcare \cite{lee2010investigation}, management \cite{carbonneau2008application}, or climate \cite{xingjian2015convolutional}.

Notwithstanding, despite gaining increasing attention, machine learning methods are still not well established in the forecasting literature, especially in the case of univariate time series. The forecasting literature is dominated by statistical methods based on linear processes, such as ARIMA \cite{chatfield2000time} or exponential smoothing \cite{gardner1985exponential}.

This matter is noticeable in the recent work by \cite{makridakis2018statistical}, where the authors present evidence that traditional statistical methods systematically outperform machine learning methods for univariate time series forecasting. This includes algorithms such as the multi-layer perceptron or Gaussian processes. 
Most of the machine learning methods tested by the authors fail to outperform a simple seasonal random walk model. Makridakis and his colleagues conclude the paper by pointing out the need to find the reasons behind the poor predictive performance shown by machine learning forecasting models relative to statistical methods. We address this question in this paper.

Our working hypothesis is that the study presented by \cite{makridakis2018statistical} is biased in one crucial aspect: sample size. The authors draw their conclusion from a large set of 1045 monthly time series used in the well-known M3 competition \cite{makridakis2000m3}. However, each of the time series is extremely small. The average, minimum, and maximum number of observations is 118, 66, and 144, respectively. We hypothesize that that these datasets are too small for machine learning models to generalize properly. 
Machine learning methods typically assume a functional form that is more flexible than that of statistical methods. Hence, they are more prone to overfit.
When the size of the data is small, the sample may not be representative of the process generating the underlying time series. In such cases, machine learning methods model the spurious behavior represented in the sample.

In this context, our goal in this paper is to compare statistical methods with machine learning methods for time series forecasting, controlling for sample size.

\subsection{Our Contribution}

To test our hypothesis, we present an empirical analysis of the impact of sample size in the relative performance of different forecasting methods. We split these methods into two categories: machine learning methods and statistical methods. Machine learning methods are often based on statistical techniques so this split is often a bit artificial. However, in the interest of consistency with previous work on this topic \cite{makridakis2018statistical}, we used the term statistical to refer to methods developed by the statistical and forecasting literature.

In our empirical analysis, we use 90 univariate time series from several domains of application. The results of our experiments show that the conclusions draw by \cite{makridakis2018statistical} are only valid when the sample size is small. That is, with small sample size, statistical methods show a better predictive performance compared to machine learning models. However, as the sample size grows, machine learning methods outperform the former.


The paper is organized as follows. In the next section, we provide a background to this paper. We formalize the time series forecasting task from a univariate perspective, and outline some state of the art methods to solve this problem. In Section \ref{sec:experiments}, we present the experiments, which are discussed in Section \ref{sec:discussion}. Finally, we conclude the paper in Section \ref{sec:conclusions}.

\section{Background}

\subsection{Time Series Forecasting}

Let $Y = \{y_1,\dots,y_n\}$ denote a time series. Forecasting denotes the process of estimating the future values of $Y$, $y_{n+h}$, where $h$ denotes the forecasting horizon. 

Quantitative approaches to time series forecasting are split into two categories: univariate and multivariate. Univariate methods refer to approaches that model future observations of a time series according to its past observations. Multivariate approaches extend univariate ones by considering additional time series that are used as explanatory variables. We will focus on univariate approaches in this work. 

The forecasting horizon is another aspect to take into account when addressing time series prediction problems. Forecasting methods usually focus on one step ahead forecasting, i.e., the prediction of the next value of a time series ($y_{n+1}$). Sometimes one is interested in predicting many steps into the future. These tasks are often referred to as multi-step forecasting \cite{taieb2012review}. Higher forecasting horizons typically lead to a more difficult predictive task due to the increased uncertainty \cite{weigend2018time}. 

\subsection{Time Series Models}

Several models for time series analysis have been proposed in the literature. These are not only devised to forecast the future behaviour of time series but also to help understand the underlying structure of the data. In this section, we outline a few of the most commonly used forecasting methods. 

The naive method, also known as the random walk forecast, predicts the future values of the time series according to the last known observation:

\begin{equation}
    \hat{y}_{n+h} = y_n
\end{equation}

\noindent There is empirical evidence that this method presents a reasonable fit for financial time series data \cite{kilian2003so}. The seasonal naive model works similarly to the naive method. The difference is that the seasonal naive approach uses the previously known value from the same season of the intended forecast:

\begin{equation}
    \hat{y}_{n+h} = y_{n+h-m}
\end{equation}

\noindent where $m$ denotes the seasonal period.

The ARMA (Auto-Regressive Moving Average) is one of the most commonly used methods to model univariate time series. ARMA(p,q) combines two components: AR(p), and MA(q).

According to the AR(p) model, the value of a given time series, $y_{n}$, can be estimated using a linear combination of the \textit{p} past observations, together with an error term $\epsilon_{n}$ and a constant term $c$ \cite{box2015time}:

\begin{equation}
    y_{n} = c + \sum^{p}_{i=1} \phi_i y_{n-i} + \epsilon_n
\end{equation}

\noindent where $\phi_i, \forall$ $i \in \{1, \dots, p\}$ denote the model parameters, and $p$ represents the order of the model. 

The AR(p) model uses the past values of the time series as explanatory variables. Similarly, the MA(q) model uses past errors as explanatory variables:
\begin{equation}
    y_{n} = \mu + \sum^{q}_{i=1} \theta_i \epsilon_{n-i} + \epsilon_n 
\end{equation}

\noindent where $\mu$ denotes the mean of the observations, $\theta_i, \forall$ $i \in \{1, \dots, q\}$ represents the parameters of the models and $q$ denotes the order of the model. Essentially, the method MA($q$) models the time series according to random errors that occurred in the past $q$ lags \cite{chatfield2000time}.

Effectively, the model ARMA(p,q) can be constructed by combining the model AR(p) with the model MA(q):
\begin{equation}
    y_{n} = c + \sum^{p}_{i=1} \phi_i y_{n-i}  + \sum^{q}_{i=1} \theta_i \epsilon_{n-i} + \epsilon_n 
\end{equation}

The ARMA(p,q) is defined for stationary data. However, many interesting phenomena in the real-world exhibit a non-stationary structure, e.g. time series with trend and seasonality. The ARIMA(p,d,q) model overcomes this limitation by including an integration parameter of order $d$. Essentially, ARIMA works by applying $d$ differencing transformations to the time series (until it becomes stationary), before applying ARMA(p,q).

The exponential smoothing model \cite{gardner1985exponential} is similar to the AR(p) model in the sense that it models the future values of time series using a linear combination of its past observations. In this case, however, exponential smoothing methods produce weighted averages of the past values, where the weight decays exponentially as the observations are older \cite{hyndman2018forecasting}. For example, in a simple exponential smoothing method, the prediction for $y_{n+1}$ can be defined as follows:

\begin{equation}
    y_{n+1} = y_{n}\beta_0 + y_{n-1}\beta_{1} + y_{n-2}\beta_{2} + \cdots
\end{equation}

\noindent where the $\{\beta_i\}$ represent the weights of past observations. There are several types of exponential smoothing methods. For a complete read, we refer to the work by \cite{hyndman2018forecasting}.

\subsection{More on the AR(p) Model}

From a machine learning perspective, time series forecasting is usually formalized as an auto-regressive task, i.e., based on an AR(p) model. This type of procedures projects a time series into a Euclidean space according to Taken's theorem regarding time delay embedding \cite{Takens1981}. 

Using common terminology in the machine learning literature, a set of observations ($\mathbf{x}_i$, $y_i$) is constructed \cite{michalski2013machine}. In each observation, the value of $y_i$ is modelled based on the past $p$ values before it: $\mathbf{x}_i = \{y_{i-1}, y_{i-2}, \dots, y_{i-p} \}$, where $y_i \in \mathbb{Y} \subset \mathbb{R}$, which represents the vector of values we want to predict, and $\textbf{x}_i \in \mathbb{X} \subset \mathbb{R}^p$ represents the feature vector. The objective is to construct a model $f : \mathbb{X} \rightarrow \mathbb{Y}$, where $f$ denotes the regression function. In other words, the principle behind this approach is to model the conditional distribution of the $i$-th value of the time series given its $p$ past values: $f$($y_{i} | \mathbf{x}_i$).
In essence, this approach leads to a multiple regression problem. The temporal dependency is modelled by having past observations as explanatory variables. Following this formulation, we can resort to any algorithm from the regression toolbox to solve the predictive task.

\subsection{Related Work}

Machine learning methods have been increasingly used to tackle univariate time series forecasting problems. However, there is a small amount of work comparing their predictive performance relative to traditional statistical methods. 
\cite{hill1996neural} compared a multi-layer perceptron with statistical methods. The neural network method is shown to perform significantly better than the latter.
\cite{ahmed2010empirical} present an analysis of different machine learning methods for this task using time series from the M3 competition \cite{makridakis2000m3}. Their results suggest that the multi-layer perceptron and Gaussian processes methods show the best predictive performance. However, the authors do not compare these methods with state of the art approaches, such as ARIMA or exponential smoothing.
In a different case study, \cite{cerqueira2019arbitrage} compare different forecasting models, including statistical methods and machine learning methods. In their analysis, the latter approaches present a better average rank (better predictive performance) relative to the former. Particularly, a rule-based regression model, which is a variant to the model tree by Quinlan, present the best average rank across 62 time series.
\cite{makridakis2018statistical} extends the study by \cite{ahmed2010empirical} by including several statistical methods in their experimental setup. Their results suggest that most of the statistical methods systematically outperform machine learning methods for univariate time series forecasting. This effect is noticeable for one-step and multi-step forecasting. The machine learning methods the authors analyze include different types of neural networks (e.g. a long short-term memory, multi-layer perceptron), the nearest neighbors method, a decision tree, support vector regression, and Gaussian processes. On the other hand, the statistical methods include ARIMA, naive, exponential smoothing, and theta, among others.

Despite the extension of their comparative study, we hypothesize that the experimental setup designed by \cite{makridakis2018statistical} is biased in terms of sample size. They use a large set of 1045 time series from the M3 competition. However, each one of these time series is extremely small in size. The average number of observations is 116. Our working hypothesis is that, in these conditions, machine learning methods are unable to learn an adequate regression function for generalization. 
In the next section, we present a new study comparing traditional statistical methods with machine learning methods. Our objective is to test the hypothesis outlined above and check whether sample size has any effect on the relative predictive performance of different types of forecasting methods.

\section{Empirical Experiments}\label{sec:experiments}

Our goal in this paper is to address the following research question: 
\begin{itemize}
    \item Is sample size important in the relative predictive performance of forecasting methods?
\end{itemize}

We are interested in comparing statistical methods with machine learning methods for univariate time series forecasting tasks. Within this predictive task we will analyse the impact of different horizons (one-step-ahead and multi-step-ahead forecasting).

\subsection{Forecasting Methods}

In this section, we outline the algorithms used for forecasting. We include five statistical methods and five machine learning algorithms.

\subsubsection{Statistical Methods}

The statistical methods used in the experiments are the following.

\begin{description}[leftmargin=*]
    \item[\texttt{ARIMA}:] The Auto-Regressive Integrated Moving Average model. We use the \textit{auto.arima} implementation provided in the forecast \textit{R} package \cite{forecast}, which controls for several time series components, including trend or seasonality;
    \item[\texttt{Naive2}] A seasonal random walk forecasting benchmark, implemented using the \textit{snaive} function available in forecast \textit{R} package \cite{forecast};
    \item[\texttt{Theta}:] The Theta method by \cite{assimakopoulos2000theta}, which is equivalent to simple exponential smoothing with drift;
    \item[\texttt{ETS}:] The exponential smoothing state-space model typically used for forecasting \cite{gardner1985exponential};
    \item[\texttt{Tbats}:] An exponential smoothing state space model with Box-Cox transformation, ARMA errors, trend and seasonal components \cite{de2011forecasting}.
\end{description}

In order to apply these models, we use the implementations available in the \textit{forecast} R package \cite{forecast}. This package automatically tunes the methods \texttt{ETS}, \texttt{Tbats}, and \texttt{ARIMA} to an optimal parameter setting. 

\subsubsection{Machine Learning Methods}

In turn, we applied the AR(p) model with the five following machine learning algorithms.

\begin{description}[leftmargin=*]
    \item[\texttt{RBR}:] A rule-based model from the \textit{Cubist} R package \cite{Cubist2014}, which is a variant of the Model Tree \cite{quinlan1993combining};
    
    \item[\texttt{RF}] A Random Forest method, which is an ensemble of decision trees \cite{breiman2001random}. We use the implementation from the \textit{ranger} R package \cite{ranger2015};
    
    \item[\texttt{GP}:] Gaussian Process regression. We use the implementation available in the \textit{kernlab} R package \cite{kernlab04};
    
    \item[\texttt{MARS}:] The multivariate adaptive regression splines \cite{friedman1991multivariate} method, using the \textit{earth} R package implementation \cite{mars2016};
    
    \item[\texttt{GLM}:] Generalized linear model \cite{mccullagh2019generalized} regression with a Gaussian distribution and a different penalty mixing. This model is implemented using the \textit{glmnet} R package  \cite{glmnet2010}.
\end{description}

These learning algorithms have been shown to present a competitive predictive performance with state of the art forecasting models \cite{cerqueira2019arbitrage}. Other widely used machine learning methods could have been included. For example, extreme gradient boosting, or recurrent neural networks. The latter have been shown to be a good fit for sequential data, which is the case of time series. Finally, we optimized the parameters of these five models using a grid search, which was carried out using validation data. The list of parameter values tested is described in Table \ref{tab:expertsspecs}. 

\begin{table}[h]
	\centering
	\caption{Summary of the learning algorithms}		
	\resizebox{.45\textwidth}{!}{%
	\begin{tabular}{llll}
	\toprule
	\textbf{ID} & \textbf{Algorithm} & \textbf{Parameter} & \textbf{Value}\\
	\midrule
	    
	    \multirow{3}{*}{\texttt{MARS}} & \multirow{3}{*}{Multivar. A. R. Splines} & Degree & \{1, 2, 3\} \\
	    
	    & & No. terms & \{2,5, 7, 15\} \\
	    
	    & & Method & \{Forward, Backward\} \\
	    
	    \midrule   
	    
        \texttt{RF} & Random forest & No. trees & \{50, 100, 250, 500\} \\
        
        \midrule   
        
        \texttt{RBR} & Rule-based regr. & No. iterations & \{1, 5, 10, 25, 50\}\\
        
	    \midrule   
	    
        \texttt{GLM} & Generalised Linear Regr. & Penalty mixing & \{0, 0.25, 0.5, 0.75, 1\}\\
        
        \midrule   
        
        \multirow{3}{*}{\texttt{GP}} & \multirow{3}{*}{Gaussian Processes} & \multirow{2}{*}{Kernel} & \{Linear, RBF,\\
	    
	    & & & Polynomial, Laplace\}\\
	    
	    & & Tolerance & \{0.001, 0.01\}\\
        
		\bottomrule    
	\end{tabular}%
	}
	\label{tab:expertsspecs}
\end{table}

\subsection{Datasets and Experimental Setup}\label{sec:es}

We centre out study in univariate time series. We use a set of time series from the benchmark database \textit{tsdl} \cite{tsdlpackage}. From this database, we selected all the univariate time series with at least 1000 observations and which have no missing values. This query returned 55 time series. These show a varying sampling frequency (daily, monthly, etc.), and are  from different domains of application (e.g. healthcare, physics, economics). For a complete description of these time series we refer to the database source \cite{tsdlpackage}. We also included 35 time series used in \cite{cerqueira2019arbitrage}. Essentially, from the set of 62 used by the authors, we selected those with at least 1000 observations and which were not originally from the \textit{tsdl} database (since these were already retrieved as described above). We refer to the work in \cite{cerqueira2019arbitrage} for a description of the time series.
In summary, our analysis is based on 90 time series. We truncated the data at 1000 observations to make all the time series have the same size.

For the machine learning methods, we set the embedding size ($p$) to 10. Notwithstanding, this parameter can be optimized using, for example, the False Nearest Neighbours method \cite{kennel1992determining}. Regarding the statistical methods, and where it is applicable, we set $p$ according to the respective implementation of  the \textit{forecast} R package \cite{forecast}.

Regarding time series pre-processing, we follow the procedure by \cite{makridakis2018statistical}. 
First, we start by applying the Box-Cox transformation to the data to stabilize the variance. The transformation parameter is optimized according to \cite{guerrero1993time}. 
Second, we account for seasonality. We consider a time series to be seasonal according to the test by \cite{wang2006characteristic}. If it is, we perform a multiplicative decomposition to remove seasonality. Similarly to \cite{makridakis2018statistical}, this process is skipped for \texttt{ARIMA} and \texttt{ETS} as they have their own automatic methods for coping with seasonality.   
Finally, we apply the Cox-Stuart test \cite{cox1955some} to determine if the trend component should be removed using first differences. This process was applied to both types of methods.

\subsection{Methodology}\label{sec:methodology}

In terms of estimation methodology, \cite{makridakis2018statistical} perform a simple holdout. 
The initial $n-18$ observations are used to fit the models. Then, the models are used to forecast the subsequent 18 observations.

We also set the forecasting horizon to 18 observations. However, since our goal is to control for sample size, we employ a different estimation procedure. Particularly, we use a prequential procedure to build a learning curve. A learning curve denotes a set of performance scores of a predictive model, in which the set is ordered as the sample size grows \cite{provost1999efficient}.
Prequential denotes an evaluation procedure in which an observation is first used to test a predictive model \cite{dawid1984present}. Then, this observation becomes part of the training set and is used to update the respective predictive model. Prequential is a commonly used evaluation methodology in data stream mining \cite{gama2010knowledge}.

We apply prequential in a growing fashion, which means that the training set grows as observations become available after testing. An alternative to this setting is a sliding approach, where older observations are discarded when new ones become available. We start applying the prequential procedure in the 18-th observation to match the forecasting horizon. In other words, the first iteration of prequential is to train the models using the initial 18 observations of a time series. These models are then used to forecast future observations. We elaborate on this approach in the next subsections. 

Following both \cite{hyndman2006another} and \cite{makridakis2018statistical}, we use the mean absolute scaled error (MASE) as evaluation metric. We also investigated the use of the symmetric mean absolute percentage error (SMAPE), but we found a large number of division by zero problems. Notwithstanding, in our main analysis, we use the rank to compare different models. A model with a rank of 1 in a particular time series means that this method was the best performing model (with lowest MASE) in that time series. We use the rank to compare the different forecasting approaches because it is a non-parametric method, hence robust to outliers.

\subsection{Results for one-step-ahead forecasting}

The first set of experiments we have carried out was designed to evaluate the impact of growing the training set size on the ability of the models to forecast the next value of the time series. To emphasize how prequential was applied, we have used the following iterative experimental procedure. In the first iteration, we have learned each model using the first 18 observations of the time series. These models were then used to make a forecast for the 19-th observation. The models were ranked according to the error of these predictions. These ranks were then average over the 90 time series. On the second iteration we grow the training set to be 19 observations and repeated the forecasting exercise this time with the goal of predicting the 20-th value. The process was repeated until we cover all available series (i.e. 1000 observations). 

\begin{figure}[t]
\centering
\includegraphics[width=.95\columnwidth]{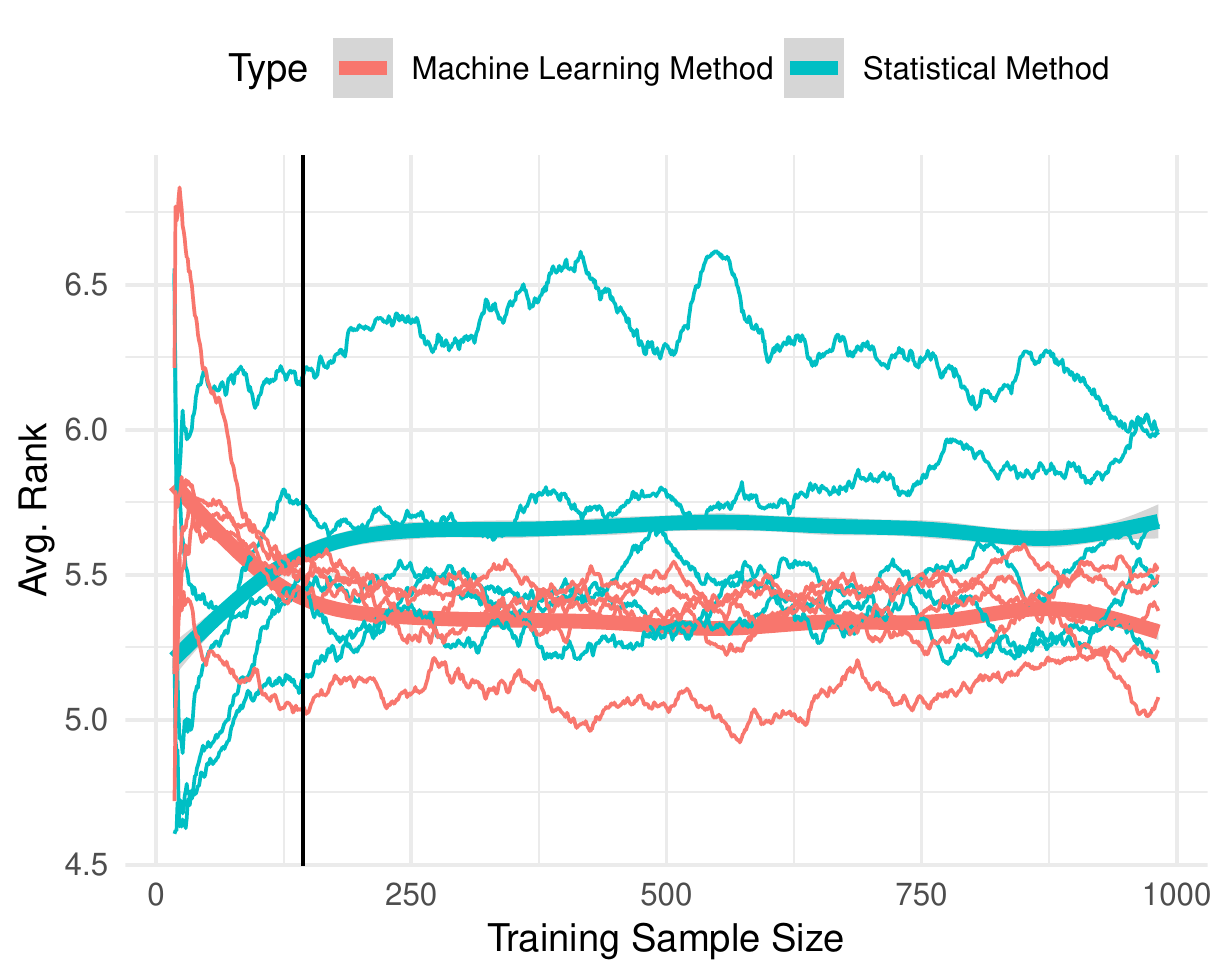} 
\caption{Learning curve using the average rank of each forecasting method, smoothed using a moving average of 50 periods. Results obtained for one-step ahead forecasting.}
\label{fig:Step1_Smooth}
\end{figure}

Figure \ref{fig:Step1_Smooth} shows the average  rank (over the 90 time series) of each model as we grow the training set according to the procedure we have just described. Particularly, this plot represents a learning curve for each forecasting model, where the models are colored by model type (machine learning or statistical). The x-axis denotes the training sample size, i.e., how much data is used to fit the forecasting models. The y-axis represents the average rank of each model across all the 90 time series, which is computed in each testing observations of the prequential procedure. For visualization purposes, we smooth the average rank of each model using a moving average over 50 observations. The two bold smoothed lines represent the smoothed average rank across each type of method according to the LOESS method. Finally, the vertical black line at point 144 represents the maximum sample size used in the experiments by \cite{makridakis2018statistical}. 

\begin{figure}[ht]
\centering
\includegraphics[width=.95\columnwidth]{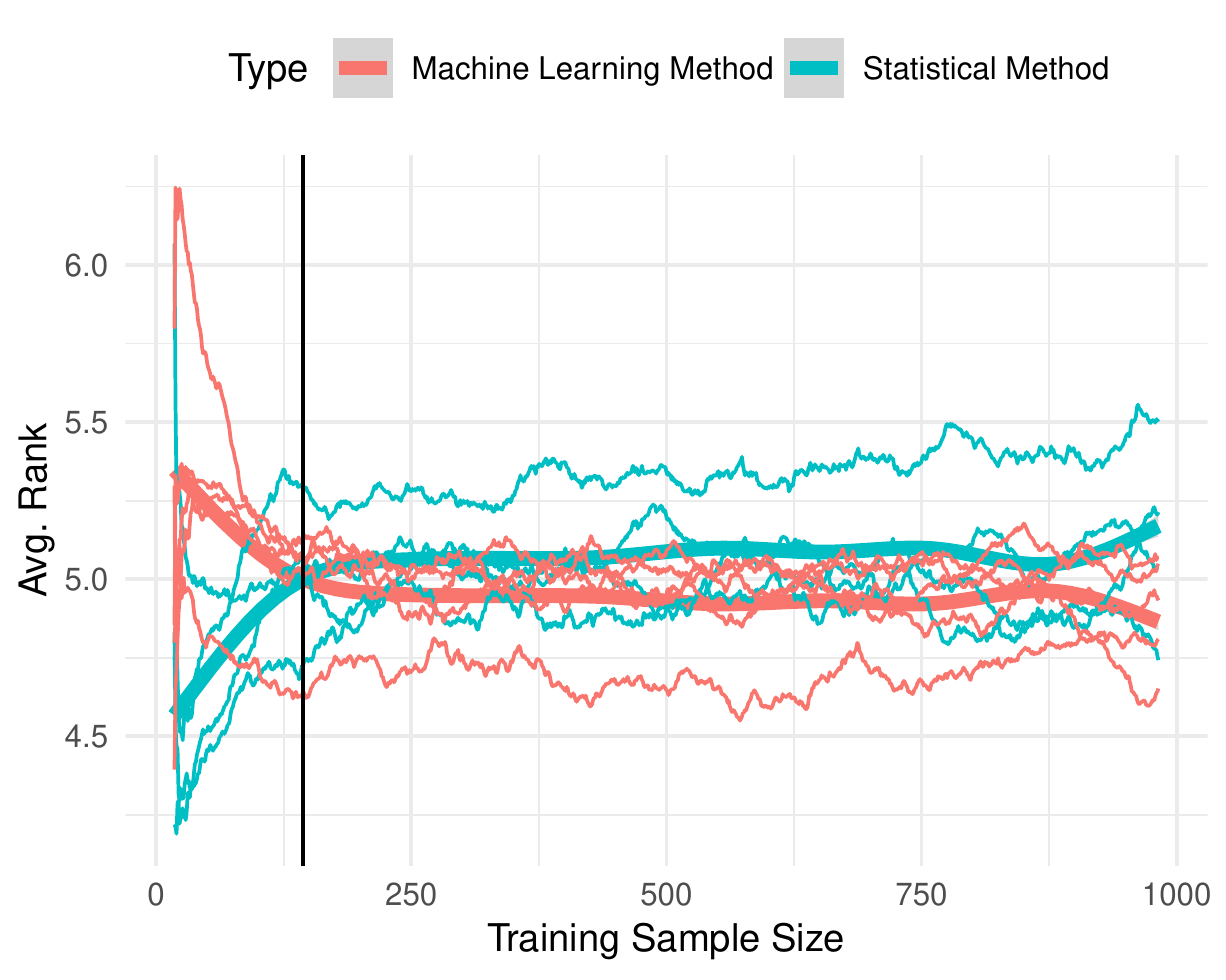} 
\caption{Learning curve using the average rank of each forecasting method (\texttt{Naive2} excluded), smoothed using a moving average of 50 periods. Results obtained for one-step ahead forecasting.}
\label{fig:Step1_Smooth_wn}
\end{figure}

\begin{figure}[!h]
\centering
\includegraphics[width=.95\columnwidth]{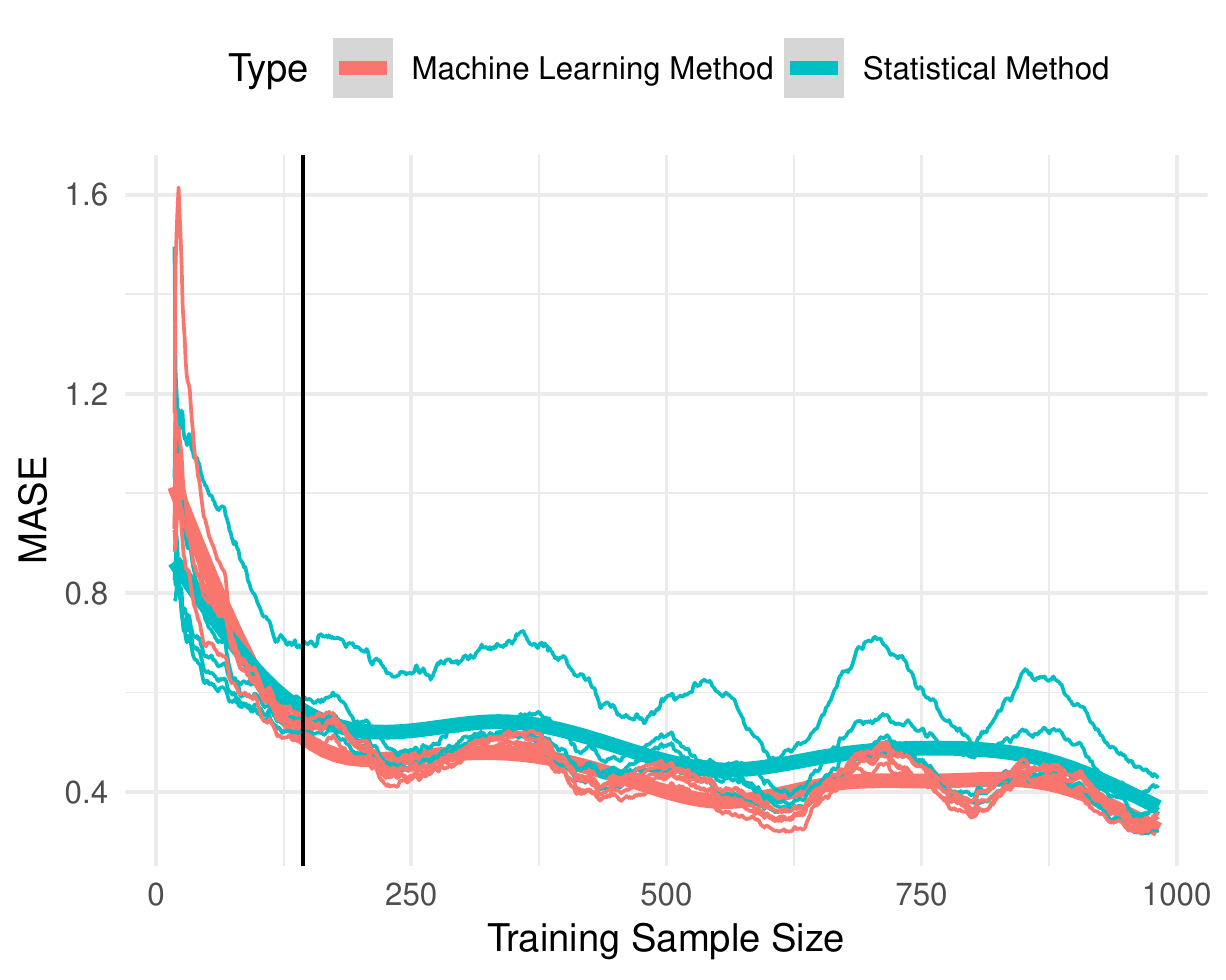} 
\caption{Learning curve using the MASE of each forecasting method, smoothed using a moving average of 50 periods. Results obtained for one-step ahead forecasting.}
\label{fig:Step1_mase}
\end{figure}

The results depicted in this figure show a clear tendency: when only few observations are available, the statistical methods present a better performance. However, as the sample size grows, machine learning methods outperform them.

Figure \ref{fig:Step1_Smooth_wn} presents a similar analysis as Figure \ref{fig:Step1_Smooth}. The difference is that we now exclude the statistical method \texttt{Naive2}. Its poor performance biased the results toward machine learning methods. Despite this, in the experiments reported by \cite{makridakis2018statistical}, this method outperforms many machine learning methods for one-step-ahead forecasting.

Our results confirm the conclusions drawn from the experiments presented by \cite{makridakis2018statistical} (the vertical black line in our figures). Namely, that statistical models outperform machine learning ones, when we only consider training sets up to 144 observations as used in that paper.

Finally, Figure \ref{fig:Step1_mase} presents a similar analysis as before using the actual MASE values. In relative terms between both types of methods, the results are consistent with the average rank analysis. This figure suggests that both types of  methods improve their MASE score as the training sample size increases.

\subsubsection{Results by Individual Model}

The average rank graphics of each model depicted in Figure \ref{fig:Step1_Smooth} show a score for each model in each of the testing observations of the prequential procedure. 
We average this value across the testing observations to determine the average of average rank of each individual model. The results of this analysis are presented in Table \ref{tbl:resultsbymodel}. The method \texttt{RBR} presents the best score.

\begin{table}[!ht]
\centering
\caption{Average of the average rank of each model across each testing observation in the prequential procedure (one-step ahead forecasting).}
\resizebox{\columnwidth}{!}{%
\begin{tabular}{rrrrrrrrrr}
  \hline
 \texttt{RBR} & \texttt{ARIMA} & \texttt{ETS} & \texttt{RF} & \texttt{GP} & \texttt{Tbats} & \texttt{MARS} & \texttt{GLM} & \texttt{Theta} & \texttt{Naive2} \\ 
  \hline
5.11 & 5.26 & 5.39 & 5.39 & 5.39 & 5.45 & 5.47 & 5.54 & 5.73 & 6.28 \\ 
   \hline
\end{tabular}%
}
\label{tbl:resultsbymodel}
\end{table}

\subsection{Results for multi-step-ahead forecasting}

In order to evaluate the multi-step ahead forecasting scenario, we used a similar approach to the one-step ahead setting. The difference is that, instead of predicting only the next value, we now predict the next 18 observations. To be precise, in the first iteration of the prequential procedure, each model is fit using the first 18 observations of the time series. These models were then used to make a forecast for the next 18 observations (from the 19-th to the 36-th). As before, the models were ranked according to the error of these predictions (quantified by MASE). On the second iteration, we grow the training set to be 19 observations, and the forecasting exercise is repeated. 

For machine learning models, we focus on an iterated (also known as recursive) approach to multi-step forecasting \cite{taieb2012review}. Initially, a model is built for one-step-ahead forecasting. To forecast observations beyond one point (i.e., $h>1$), the model uses its predictions as input variables in an iterative way. We refer to the work by \cite{taieb2012review} for an overview and analysis of different approaches to multi-step forecasting. 

\begin{figure}[h]
\centering
\includegraphics[width=\columnwidth]{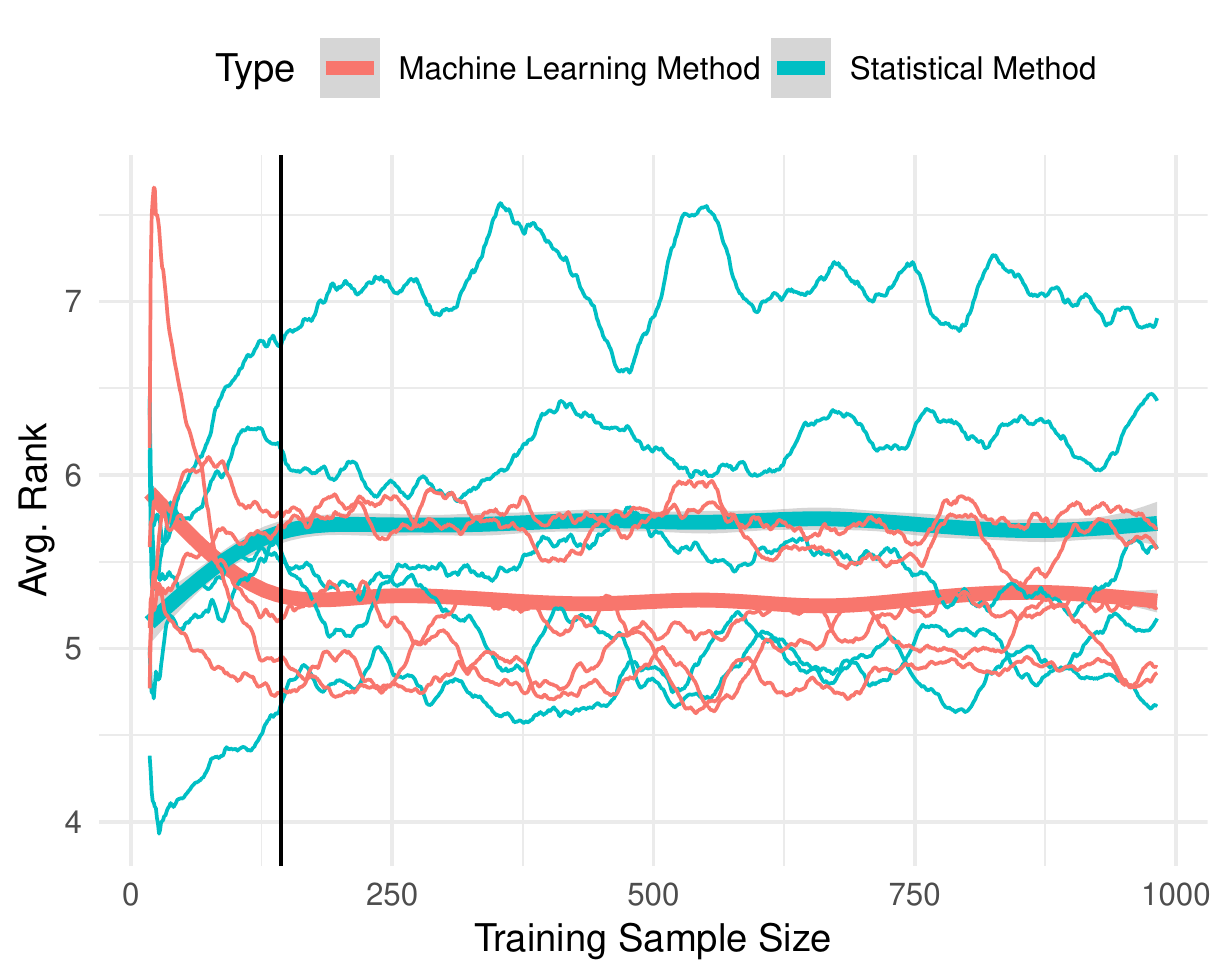}
\caption{Learning curve using the smoothed average rank of each forecasting method for $h=18$. Results for multi-step ahead forecasting.}
\label{fig:Step18_ALL}
\end{figure}

\begin{figure}[!bh]
\centering
\includegraphics[width=\columnwidth]{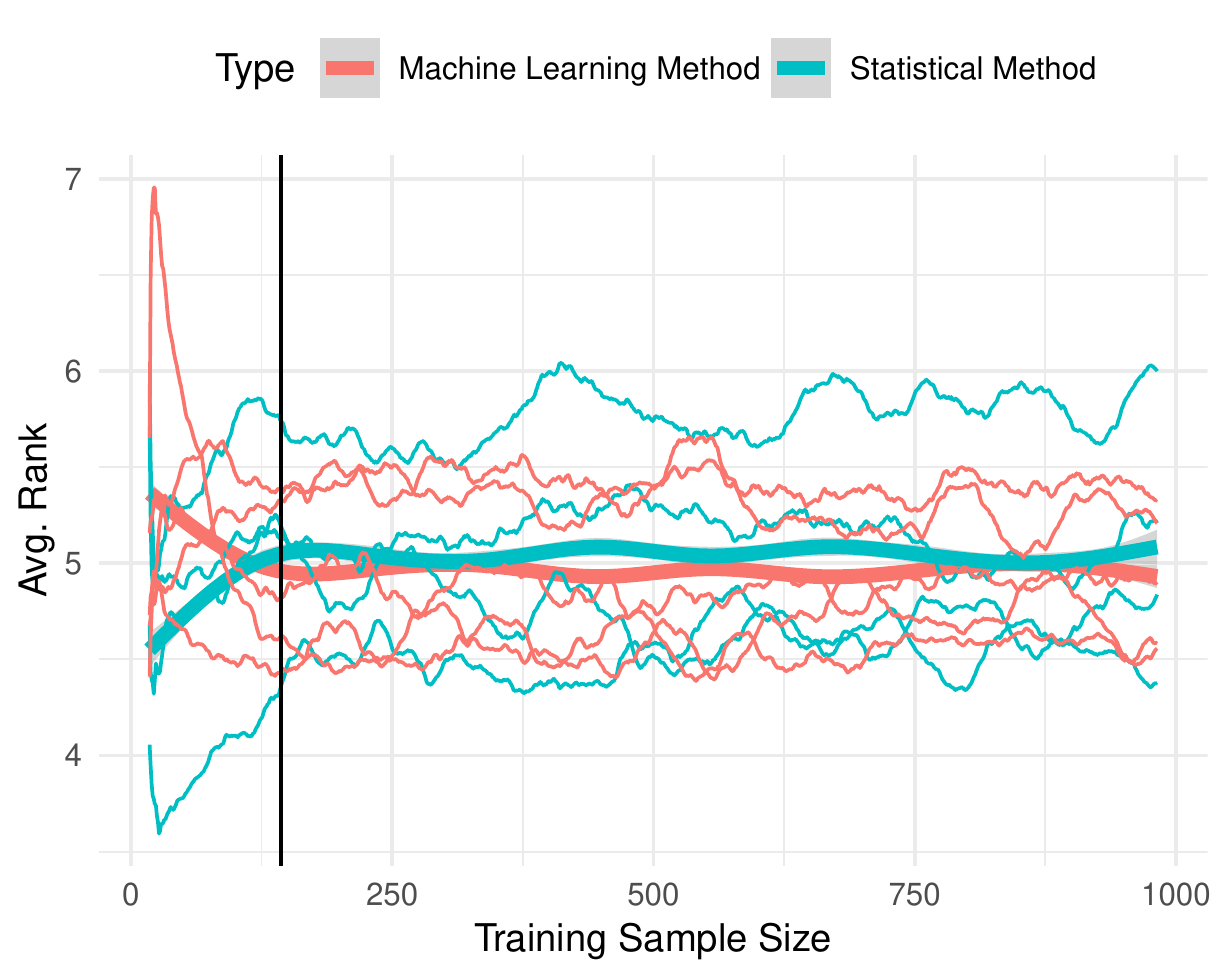}
\caption{Learning curve using the smoothed average rank of each forecasting method for $h=18$, \texttt{Naive2} excluded.}
\label{fig:Step18_noNaive}
\end{figure}

\begin{figure}[!th]
\centering
\includegraphics[width=\columnwidth]{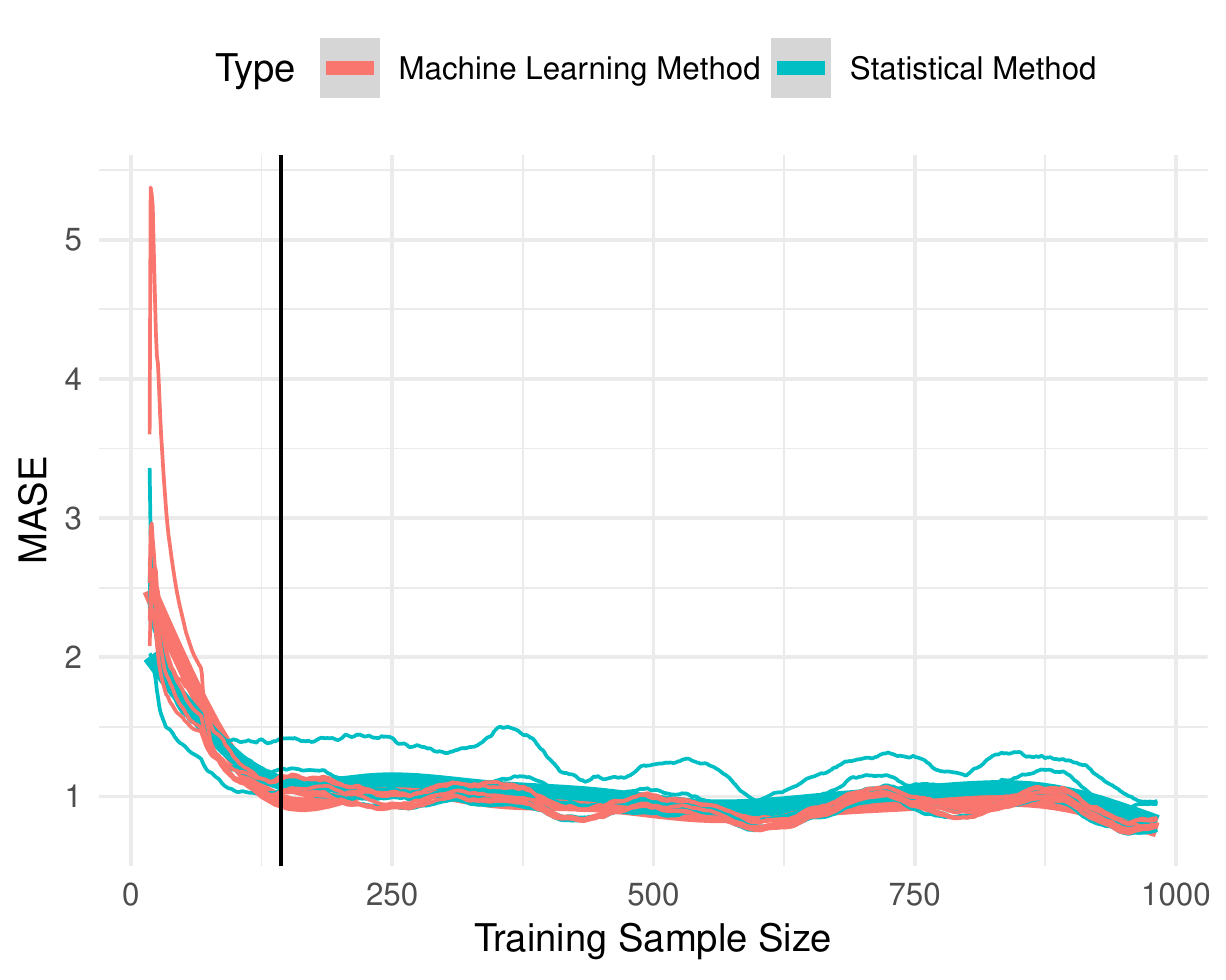}
\caption{Learning curve using the MASE of each forecasting method and for $h=18$, smoothed using a moving average of 50 periods.}
\label{fig:Step18_mase}
\end{figure}

Figures \ref{fig:Step18_ALL} and \ref{fig:Step18_noNaive} show the average (over the 90 time series) rank of each model as we grow the training set according to the procedure we have just described. The figures are similar, but the results of the \texttt{Naive2} method were excluded from the second one using the same rationale as before. 

Analyzing the results by model type, the main conclusion in this scenario is similar to the one obtained in the one-step ahead setting: according to the smoothed lines, statistical methods are only better than machine learning ones when the sample size is small. In this setting, however, the two types of methods seem to even out as the training sample size grows (provided we ignore \texttt{Naive2}). According to Figure \ref{fig:Step18_mase}, the MASE score of the models in this scenario are considerably higher than that of the one-step ahead forecasting setting. This is expected given the underlying increased uncertainty.

\begin{table}[ht]
\centering
\caption{Average of the average rank of each model across each testing observation in the prequential procedure (multi-step ahead forecasting).}
\resizebox{\columnwidth}{!}{%
\begin{tabular}{rrrrrrrrrr}
  \hline
 \texttt{ARIMA} & \texttt{RBR} & \texttt{GLM} & \texttt{Tbats} & \texttt{GP} & \texttt{ETS} & \texttt{MARS} & \texttt{RF} & \texttt{Theta} & \texttt{Naive2} \\ 
  \hline
4.77 & 4.95 & 4.97 & 5.11 & 5.24 & 5.43 & 5.69 & 5.74 & 6.12 & 6.98 \\ 
   \hline
\end{tabular}%
}
\label{tbl:resultsbymodelMS}
\end{table}

Table \ref{tbl:resultsbymodelMS} presents the results by individual model in a similar manner to Table \ref{tbl:resultsbymodel} but for multi-step forecasting. In this setting, the model with best score is \texttt{ARIMA}.

\subsection{Computational Complexity}

In the interest of completeness, we also include an analysis of the computational complexity of each method. We evaluate this according to the computational time spent by a model, which we define as the time a model takes to complete the prequential procedure outlined in Section \ref{sec:methodology}.  Similarly to \cite{makridakis2018statistical}, we define the computational complexity (CC) of a model $m$ as follows:
\begin{equation}
    \text{CC} = \frac{\text{Computational Time } m}{\text{Computational Time } \texttt{Naive2}}
\end{equation}

\noindent Essentially, we normalize the computational time of each method by the computational time of the \texttt{Naive2} method.

\begin{figure}[ht]
\centering
\includegraphics[width=\columnwidth]{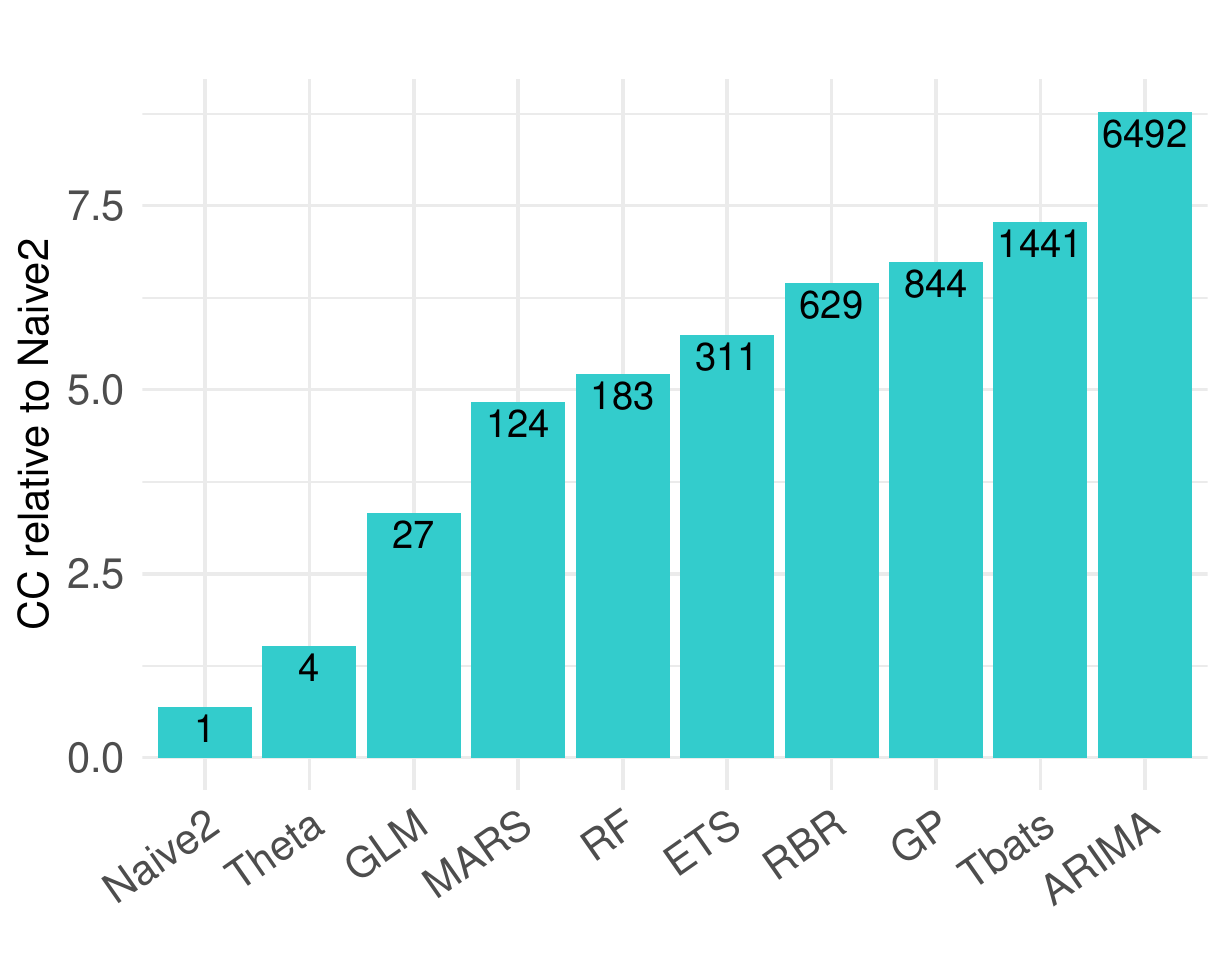}
\caption{Computational complexity of each method relative to the \texttt{Naive2} benchmark model (log scaled)} 
\label{fig:cc}
\end{figure}

The results are shown in Figure \ref{fig:cc} as a bar plot. The bars in the graphic are log scaled. The original value before taking the logarithm is shown within each bar. 

From the figure, the method with worse CC is \texttt{ARIMA}, followed by \texttt{Tbats}. The results are driven by the fact that the implementations of the statistical methods (except \texttt{Naive2} and \texttt{Theta}) include comprehensive automatic parameter optimization in the \textit{forecast} R package. The optimization of machine learning methods carried out in our experiments was not as exhaustive. Therefore, the CC of these methods is not as high as the one shown by \texttt{ARIMA} and \texttt{Tbats}.

\section{Discussion}\label{sec:discussion}

In the previous section, we carried a simple experiment to show that sample size matters when comparing machine learning methods with statistical methods for univariate time series forecasting. In this section, we discuss these results.

\subsection{Experimental Setup}\label{sec:disc_es}

We applied the statistical methods using the widely used R package \textit{forecast} \cite{forecast}. This package contains implementations that search for the best parameter settings of the models, and is considered a software package for automated time series forecasting models \cite{taylor2018forecasting}. 
Regarding the machine learning models, we focused on building AR(p) models using these learning algorithms. Other approaches could be analyzed, for example coupling an AR(p) model with a recurrent architecture \cite{dietterich2002machine} or with summary statistics.

Our choice of machine learning algorithms was driven by a recent study \cite{cerqueira2019arbitrage}, but other learning algorithms could be applied. For example, we did not apply any neural network, which have been successfully applied to sequential data problems \cite{chung2014empirical}. Our goal was to show that machine learning in general is a valid approach to time series forecasting, even without an extensive model selection procedure nor a parameter tuning process as extensive as the one used by the \textit{forecast} package.

\subsection{One-step vs Multi-step Forecasting}

The results obtained from the multi-step forecasting scenario are different from those obtained from the one-step ahead scenario. Particularly, machine learning methods do not gain the upper hand in terms of predictive performance when we increase the sample size. 

Some possible reasons are the following: since multi-step forecasting represents a task with an increased uncertainty \cite{weigend2018time}, machine learning methods may need more data to cope with this issue -- in future work, we will continue the learning curve to assess this possibility; a different approach to multi-step forecasting may be better -- we focused on an iterative approach, but other possibilities are available in the literature \cite{taieb2012review}. 

\subsection{On Sample Size and the \textit{No Free Lunch} Theorem}

Our main claim in this work is that sample size matters when comparing different forecasting models. We backed this claim using 90 time series comprised of 1000 observations. We do not claim that this number is the optimal sample size for fitting forecasting models. It is difficult to tell apriori the amount of data necessary to solve a predictive task. It depends on many factors, such as the \textit{complexity} of the problem  or the \textit{complexity} of the learning algorithm. 

We believe that the work by \cite{makridakis2018statistical} is biased towards small, low frequency, datasets. Naturally, the evidence that machine learning models are unable to generalize from small datasets can be regarded as a limitation relative to traditional statistical ones. However, machine learning can make an important impact in larger time series. Technological advances such as the widespread adoption of sensor data enabled the collection of large, high frequency, time series. This occurrence lead to new forecasting problems, where machine learning models have been shown to be useful \cite{voyant2017machine,xingjian2015convolutional}. 

Finally, we also remark that, even with a large amount of data, it is not obvious that a machine learning method would outperform a statistical method. This reasoning is in accordance with the \textit{No Free Lunch} theorem \cite{wolpert1996lack}, which states that no learning algorithm is the most appropriate in all scenarios. The same rationale can be applied to small datasets.

\section{Final Remarks}\label{sec:conclusions}

Makridakis claims that machine learning practitioners ``working on forecasting applications need to do something to improve the accuracy of their methods'' \cite{makridakis2018statistical}. We claim that these practitioners should start by collecting as much data as possible. Moreover, it is also advisable for practitioners to include both types of forecasting methods in their studies to enrich the experimental setup. 

The code to reproduce the experiments carried out in this paper can be found at \url{https://github.com/vcerqueira/MLforForecasting}.

\section*{Acknowledgements}

The work of V. Cerqueira was financially supported by \textit{Fundação para a Ciência e a Tecnologia} (FCT), the Portuguese funding agency that supports science, technology, and innovation, through the Ph.D. grant SFRH/BD/135705/2018. The work of L. Torgo was undertaken, in part, thanks to funding from the Canada Research Chairs program.

\end{document}